\documentclass{article} 
\usepackage{arxiv_ausesol,times}

\usepackage{amsmath,amsfonts,bm}

\usepackage{mathtools} 
\usepackage{amssymb}
\usepackage{amsthm}
\usepackage{enumitem}
\usepackage{multirow}
\usepackage{siunitx}
\usepackage{centernot}
\usepackage{subcaption}

\def\eqref#1{equation~\ref{#1}}

\def\1{\bm{1}}

\DeclareMathAlphabet{\mathsfit}{\encodingdefault}{\sfdefault}{m}{sl}
\SetMathAlphabet{\mathsfit}{bold}{\encodingdefault}{\sfdefault}{bx}{n}

\usepackage{euscript}

\usepackage{multicol}
\usepackage{booktabs}
\usepackage{makecell}
\usepackage{tabstackengine}
\stackMath
\setstackgap{L}{1.2\baselineskip}
\fixTABwidth{T}

\makeatletter
\newcommand{\raisemath}[1]{\mathpalette{\raisem@th{#1}}}
\newcommand{\raisem@th}[3]{\raisebox{#1}{$#2#3$}}
\makeatother

\newcommand*\restr[2]{{
  #1 
  \kern-\nulldelimiterspace 
  \left.\kern-\nulldelimiterspace 
  \vphantom{\big|} 
  \right|_{\raisemath{1pt}{#2}} 
  }}

\usepackage{thmtools, thm-restate}
\theoremstyle{plain}

\usepackage{hyperref}
\usepackage{url}

\makeatletter
\renewcommand{\headrulewidth}{0pt} 
\makeatother

\title{Physics-Informed Learning of Flow Distribution and Receiver Heat Losses in Parabolic Trough Solar Fields}

\author{%
  Stefan Matthes \\
  fortiss GmbH \\
  Munich, BY 80805, Germany \\
  \texttt{matthes@fortiss.org} \\
\And
Markus Schramm\\
CSP Services España S.L. \\
Almería, 04001, Spain \\
\texttt{m.schramm@cspservices.de} \\
}

\iclrfinalcopy 
\begin{document}

\maketitle

\begin{abstract}
Parabolic trough Concentrating Solar Power (CSP) plants operate large hydraulic networks of collector loops that must deliver a uniform outlet temperature despite spatially heterogeneous optical performance, heat losses, and pressure drops. While loop temperatures are measured, loop-level mass flows and receiver heat-loss parameters are unobserved, making it impossible to diagnose hydraulic imbalances or receiver degradation using standard monitoring tools. 

We present a physics-informed learning framework that infers (i) loop-level mass-flow ratios and (ii) time-varying receiver heat-transfer coefficients directly from routine operational data. The method exploits nocturnal homogenization periods—when hot oil is circulated through a non-irradiated field—to isolate hydraulic and thermal-loss effects. A differentiable conjugate heat-transfer model is discretized and embedded into an end-to-end learning pipeline optimized using historical plant data from the 50\,MW Andasol 3 solar field.


The model accurately reconstructs loop temperatures (RMSE $<2^\circ$C) and produces physically meaningful estimates of loop imbalances and receiver heat losses. Comparison against drone-based infrared thermography (QScan) shows strong correspondence, correctly identifying all areas with high-loss receivers. This demonstrates that noisy real-world CSP operational data contain enough information to recover latent physical parameters when combined with appropriate modeling and differentiable optimization.
 
\end{abstract}

\section{Introduction}

Parabolic trough power plants represent the most widely deployed configuration within Concentrating Solar Power (CSP) technology. In these systems, long rows of parabolic mirrors track the sun and concentrate direct normal irradiance (DNI) onto an absorber tube through which a heat-transfer fluid (HTF), typically a synthetic oil, circulates. The absorbed thermal energy is transferred to a power block or a thermal storage system, enabling dispatchable electricity generation. A solar field is composed of numerous collector loops, each consisting of several Solar Collector Assemblies (SCAs) connected in series.

\begin{figure}[ht]
    \centering

    \begin{subfigure}{0.48\linewidth}
        \centering
        \includegraphics[width=\linewidth]{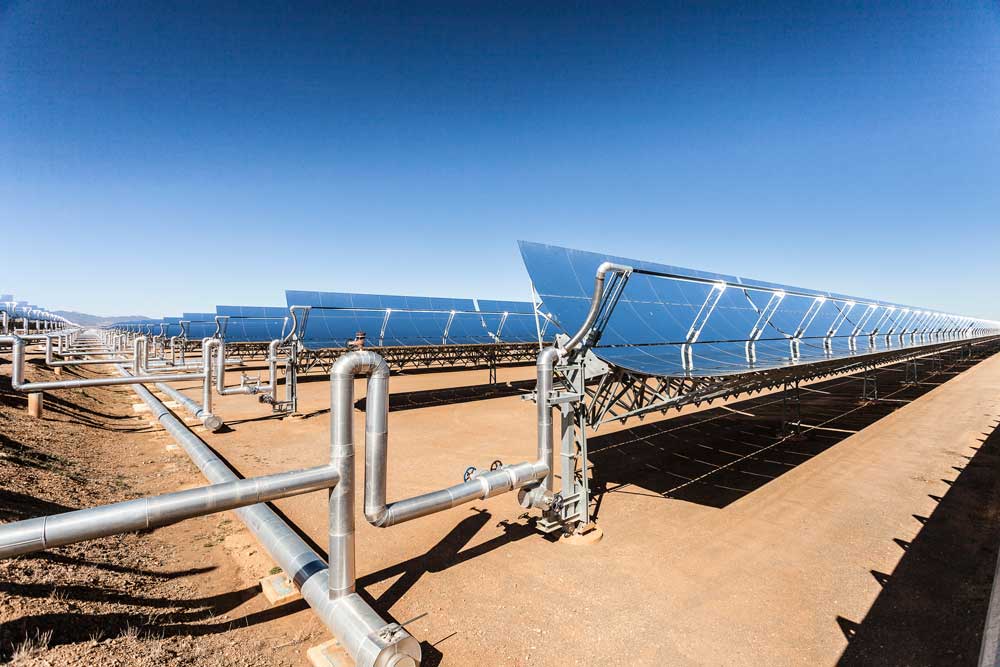}
    \end{subfigure}
    \hfill
    \raisebox{4mm}{%
    \begin{subfigure}{0.48\linewidth}
        \centering
        \includegraphics[width=\linewidth]{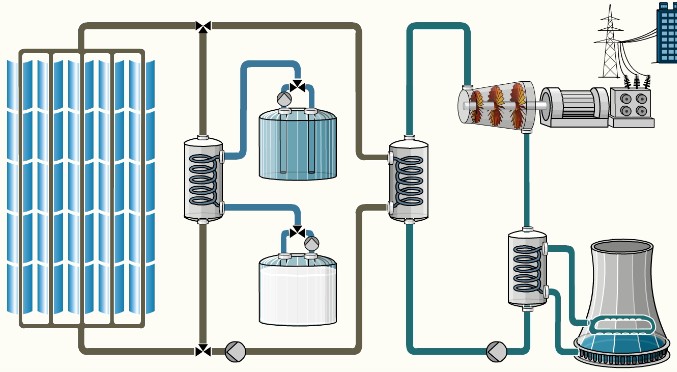}
    \end{subfigure}
    }
    \caption{Left: Image of solar collector assemblies connected header pipes. Right: Schematic overview of a parabolic trough solar power plant. Source: https://marquesadosolar.com/}
\end{figure}

The plant studied in this work, Andasol 3 in southern Spain, is representative of modern large-scale CSP installations. Its solar field comprises 152 loops grouped into four subfields, with five temperature sensors per loop (four in the center of each SCAs and one at the outlet).
The fundamental operational challenge remains that the HTF delivered by the outlet of the loops must meet strict temperature conditions. Therefore, each loop requires a mass flow that reflects its own optical efficiency and thermal losses to achieve the common target outlet temperature. Disbalances in the hydraulic network lead to losses due to a lower temperature rise or forced de-focusing of collectors.  

Therefore, operators adjust loop valves to achieve this setpoint, but the valve position is not a reliable indicator of loop efficiencies: it also reflects the loop’s location within the hydraulic network with it's local pressure drop and piping details. Critically, loop-level mass flow is not measured, so the apparent success of holding outlet temperature does not reveal whether a loop is under-performing nor disentanglement of underlying mechanisms is possible. As a result, the field diagnostics remain ambiguous.

Modern Concentrated Solar Power (CSP) plants like Andasol 3 generate enormous streams of operational data, creating opportunities for data-driven condition monitoring. By leveraging existing plant data (temperatures, irradiance, valve positions, etc.), ML-based analytics can uncover subtle patterns indicative of faults or efficiency degradation. 

Recent research on data-driven condition monitoring in solar thermal systems spans both anomaly-detection methods and supervised fault-diagnosis approaches. Early work by \citet{Maciejewski2008} applied data-driven techniques for condition monitoring in a direct steam generation plant. Using principal component analysis and partial least squares to reduce the high-dimensional sensor data to two latent variables, the authors could distinguish normal and abnormal operating states, including detecting a pump malfunction. The study demonstrated that simple latent-variable methods could effectively highlight anomalies for the used data. A further early example is the ANN-based fault detection system by \citet{KALOGIROU2008164} for solar water heaters. Separate models were trained for each temperature sensor using fault-free TRNSYS simulations, and real measurements were compared against these predictions. Residuals fed into a simple diagnosis module allowed classification into three fault types. As only selected temperature sensors were monitored, the method could detect only faults that produced observable temperature deviations.

Simulation-based fault detection has also been explored in low-temperature solar thermal systems. \citet{DEKEIZER2013} presented a fault detection method that compares measured operation with TRNSYS simulations. Expected value ranges for selected features are derived from sensor uncertainty bounds; measurements outside these ranges trigger “symptoms,” which are then manually interpreted to identify faults. The method demonstrated capabilities in detecting low-yield conditions. In supervised CSP-focused work, \citet{munoz2019} developed a condition-based maintenance tool for a parabolic trough field in Chile. Their system employs a suite of four machine learning models to detect and classify specific faults. In tests, this multi-model approach could correctly identify the source of a fault about $80\%$ of the time (for faults inducing $\ge 20\%$ performance loss in the field) when the issue lay in the solar field or heat exchanger.

More recently, data-driven fault detection systems have expanded to large solar thermal installations. \citet{FEIERL2023} presented Fault-Detective, a data-driven fault detection method for large solar thermal systems. It identifies correlated sensors, trains Random Forest models to predict target variables, and raises alarms when prediction errors exceed defined thresholds. In tests on three installations, the method performed well for thermal-power monitoring but produced many false alarms for temperature signals, mainly due to anomalies such as consecutive days with bad weather being flagged as faults.

A distinct line of work has emerged within CSP itself, including approaches targeted specifically at parabolic trough collectors. \citet{braun2023} proposed an unsupervised anomaly-detection method for parabolic trough solar fields based on multivariate time-series segmentation and density-based outlier scoring. Operational loop-level data from one year are segmented—using either window-sliding or periodic schemes—and transformed into simple statistical features. Anomalies are then identified via the Local Outlier Factor (LOF) algorithm, which quantifies how strongly each segment deviates from normal operational patterns. Although the two segmentation strategies flag different time points, they highlight similar underlying causes of anomalies and reveal loops with consistently elevated outlierness. 

Deep learning–based fault diagnosis for parabolic trough collectors has been advanced in a series of works by Ruiz-Moreno and co-authors. \citet{ruizmoreno2022_2} first introduced a deep-learning–based methodology for fault detection and isolation in parabolic-trough collectors, using a hierarchical three-layer scheme that combines a multilayer perceptron with two decoupling stages for flow-rate and thermal-loss faults. Applied to simulations of the ACUREX plant, this approach achieved fault-classification accuracies above $80\%$ and over $90\%$ when all three layers were used in combination. Building on this work, \citet{ruizmoreno2023} later proposed an ANN-based approach for fault detection and isolation in a simulated 50 MW parabolic trough plant. A feedforward network was trained on synthetic data produced by a dynamic plant model, enabling it to recognize deviations from normal operation. Their results show that detection accuracy improved from roughly $72\%$ to $91\%$ after optimizing the input features and the fault ranges used for training.

Prior literature demonstrates the potential of data-driven methods for detecting abnormal behavior in solar thermal systems, yet these approaches generally lack embedded physics, making it difficult to attribute anomalies to root causes. Physics-informed models, by contrast, provide interpretability and allow the recovery of latent quantities such as flow distribution or receiver heat losses.

This work proposes a physics-informed learning framework that turns nighttime homogenization sequences into a natural experiment allowing the extraction of hydraulic and thermal parameters. We show that the method recovers physically meaningful quantities and agrees strongly with drone-based infrared measurements.

\section{Approach}

At night or before dawn, the solar field undergoes \emph{homogenization}: hot HTF is circulated to thermally equalize it before daytime operation. During these intervals:
\begin{itemize}
    \item there is no solar irradiation;
    \item mirror soiling, tracking, misalignment, and torsion are irrelevant;
    \item only \emph{hydraulic effects} and \emph{heat losses} drive the temperature evolution.
\end{itemize}
Thus, measured temperatures encode (i) the loop mass-flow distribution and (ii) receiver heat-loss conditions—exactly the latent variables we seek.

\subsection{Conjugate Heat-Transfer Model}
\begin{figure}[ht]
    \centering
    \includegraphics[width=0.9\linewidth]{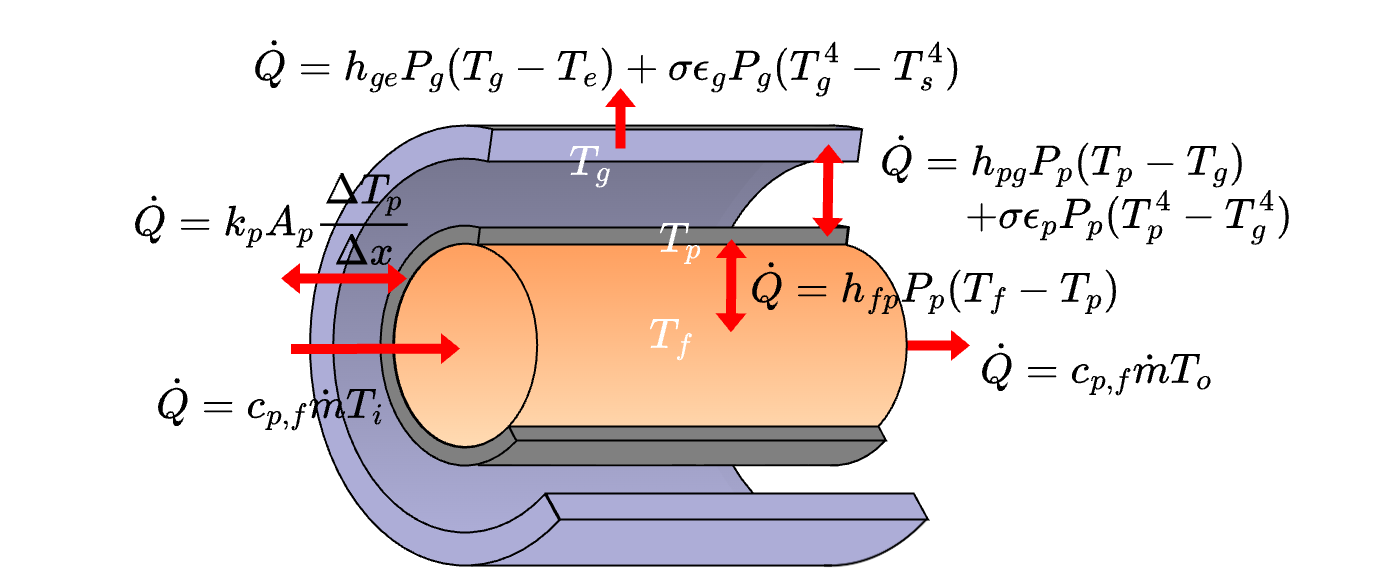}
    \caption{Considered heat flows in the receiver control volume.}
    \label{fig:conjugate_heat_transfer}
\end{figure}

Figure~\ref{fig:conjugate_heat_transfer} illustrates the heat flows: convection between HTF and pipe, conduction along the steel, radiative and convective exchange between pipe and glass envelope, and envelope losses to ambient. Axial conduction in the fluid and envelope is neglected.

The governing equations are:
\begin{align}
\frac{\partial T_f}{\partial t} &= -u \frac{\partial T_f}{\partial x} 
    + \frac{h_{fp} P_p}{c_{v,f} A_f} (T_p - T_f), \\
(c_{v,p} A_p)\frac{\partial T_p}{\partial t} &=
    - h_{fp} P_p (T_p - T_f)
    + h_{pg}^t P_p (T_g - T_p)
    + \epsilon_p \sigma P_p (T_g^4 - T_p^4)
    + k_p A_p \frac{\partial^2 T_p}{\partial x^2}, \\
(c_{v,g} A_g)\frac{\partial T_g}{\partial t} &=
    - h_{pg}^t P_p (T_g - T_p)
    + h_{ge} P_g (T_e - T_g)
    + \epsilon_g \sigma P_g (T_s^4 - T_g^4).
\end{align}

Parameter descriptions are summarized in Table~\ref{tbl:parameters}.
Here $h_{pg}^t$ is the \emph{vacuum-dependent} pipe--glass heat-transfer coefficient and is learned per period.
The nightly sky temperature, $T_s$, is assumed to be 20 °C below the ambient temperature $T_e$.

\begin{table}[ht]
\centering
\footnotesize
\begin{tabular}{llp{8.5cm}}
\hline
Symbol & Units & Description \\ \hline
$T_f, T_p, T_g$ & K & Fluid, pipe, and glass temperatures \\
$T_e, T_s$ & K & Ambient temperature, sky temperature \\
$T_{\mu}$ & K & Average fluid temperature in loop \\
$u$ & m/s & Fluid velocity \\
$A_f, A_p, A_g$ & m$^2$ & Cross-sectional areas of fluid, pipe, and glass \\
$P_p, P_g$ & m & Wetted perimeters of pipe and glass \\
$c_{v,f}, c_{v,p}, c_{v,g}$ & J/(kg\,K) & Volumetric heat capacities \\
$h_{fp}, h_{ge}$ & W/(m$^2$K) & Fluid--pipe and glass--environment convective coefficient \\
$h_{pg}^t$ & W/(m$^2$K) & Pipe--glass heat-transfer coefficient (vacuum dependent, learned) \\
$k_p$ & W/(m\,K) & Pipe axial conductivity \\
$\epsilon_p, \epsilon_g$ & -- & Pipe and glass emissivities \\
$\sigma$ & W/(m$^2$K$^4$) & Stefan--Boltzmann constant \\
$\dot{V}_h$ & m$^3$/s & Measured subfield volume flow \\
$\dot{m}_h$ & kg/s & Subfield mass flow \\
$\dot{m}_i$ & kg/s & Mass flow in loop $i$ (inferred) \\
$\beta_i$ & -- & Mass-flow ratio for the $i$-th loop \\
$\rho_f$ & kg/m$^3$ & Fluid density \\
$\omega^t$ & -- & Valve-state latent vector (learned)\\
$a, b$ & -- & Linear scaling coefficients for valve model (learned)\\
$\alpha$ & -- & Flow-bias correction factor (learned)\\
\hline
\end{tabular}
\caption{Model parameters and descriptions.}
\label{tbl:parameters}
\end{table}

\subsection{Mass-Flow Allocation Model}

Only the total subfield volume flow $\dot{V}_h$ is measured, so individual loop flows must be inferred. We define
\[
\dot{m}_i = \beta_i\,\dot{m}_h,
\]
where we model the loop flow ratio of the $i$-th loop by
\[
\beta_i = \mathrm{softmax}_i\!\left((a T_\mu + b)\circ \omega^t \right).
\]
Here, $T_\mu$ is a vector containing the average loop temperatures for all loops in a subfield, $a$ and $b$ are learnable constants, $\circ$ denotes the Hadamard product (or element-wise product), and $\omega^t$ is a learnable valve state vector that only changes when operators adjust the valves.

The loop velocity follows from continuity:
\[
u = \beta_i(T_\mu) 
\frac{\rho_f(T_h)}{\rho_f(T_f)}
\frac{\alpha\, \dot{V}_h}{A_f},
\]
with $\alpha$ a learned volumetric-flow bias correcting measurement drift.

\newpage
\subsection{Parameters}

We distinguish:
\begin{itemize}
    \item \textbf{Known constants}: geometry, emissivities, thermal capacities, pipe conductivity, and fluid properties.
    \item \textbf{Global learnable parameters}: $a, b, \alpha$.
    \item \textbf{Time-varying learnable parameters}: $h_{pg}^t$, $\omega^t$.
\end{itemize}

All equations are discretized using forward Euler (5\,s timestep, $\sim$10\,m spatial segments). The entire model is differentiable and optimized via Stochastic Gradient Descent (SGD) over 176 homogenization sequences spanning one year.

\section{Evaluation}

\begin{figure}[ht]
    \centering
    \includegraphics[width=0.8\linewidth]{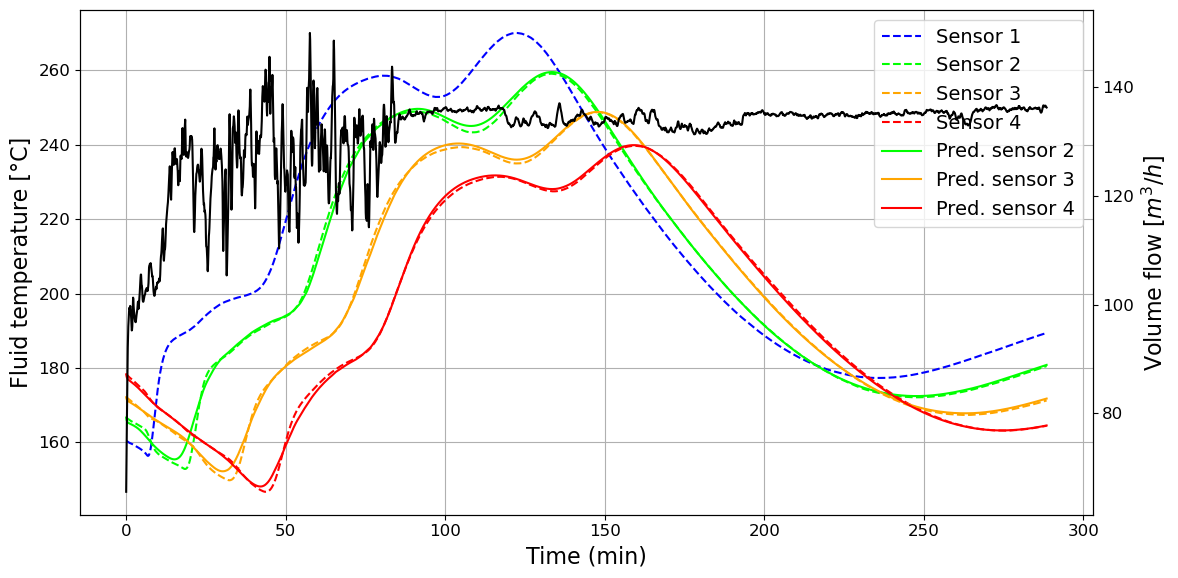}
    \caption{Measured and predicted temperatures along a single loop. The black line denotes the subfield volume flow.}
    \label{fig:temperature_prediction}
\end{figure}

Figure~\ref{fig:temperature_prediction} shows measured vs.\ predicted temperatures for a validation sequence where hotter heat transfer fluid passes through the temperature sensors of a loop. The overall RMSE remains below $2^\circ$C, demonstrating that the physics-informed model captures the dominating dynamics of nighttime heat exchange.

\begin{figure}[ht]
    \centering
    \begin{subfigure}{0.58\linewidth}
        \centering
        \includegraphics[width=\linewidth]{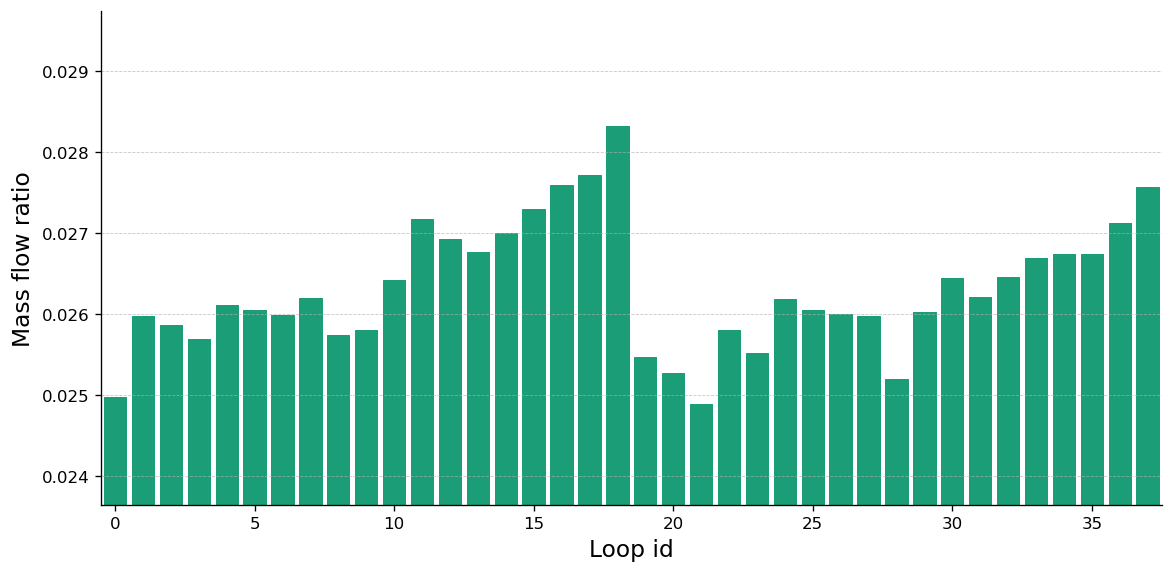}
    \end{subfigure}
    \hfill
    \begin{subfigure}{0.4\linewidth}
        \centering
        \includegraphics[width=\linewidth]{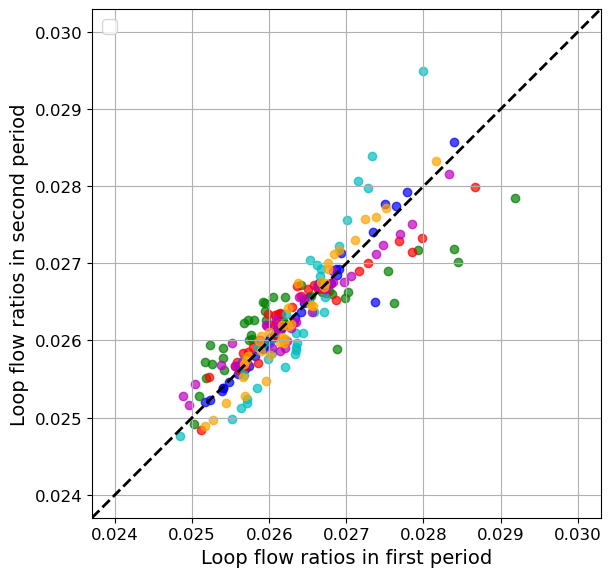}
    \end{subfigure}
    \caption{Left: Predicted mass-flow ratios for loops in a subfield. Right: Self-consistency comparison across six pairs of periods with identical valve states (indicated by different colors) and an average coefficient of determination of 0.78.}
    \label{fig:balancing}
\end{figure}

Figure~\ref{fig:balancing} illustrates the inferred mass-flow ratios in a subfield. Because no ground truth exists, we perform a self-consistency check: we optimize $\omega^t$ separately for consecutive periods with identical valve settings. Consistent predictions validate the identifiability of the mass-flow parameters.

\begin{figure}[ht]
    \centering
    \raisebox{5mm}{%
    \begin{subfigure}{0.37\linewidth}
        \centering
        \includegraphics[width=\linewidth]{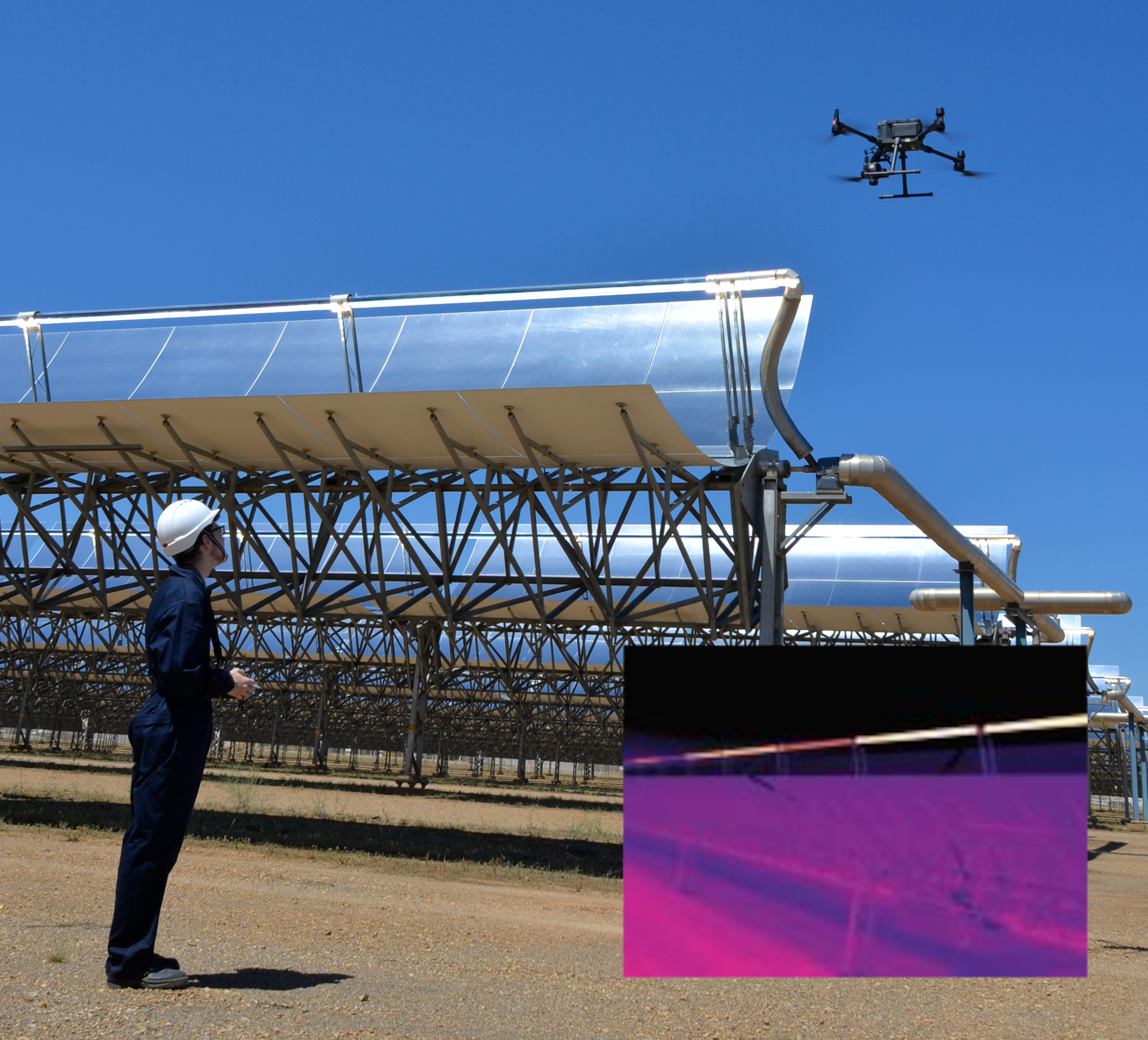}
    \end{subfigure}
    }
    \hfill
    \begin{subfigure}{0.6\linewidth}
        \centering
        \includegraphics[width=\linewidth]{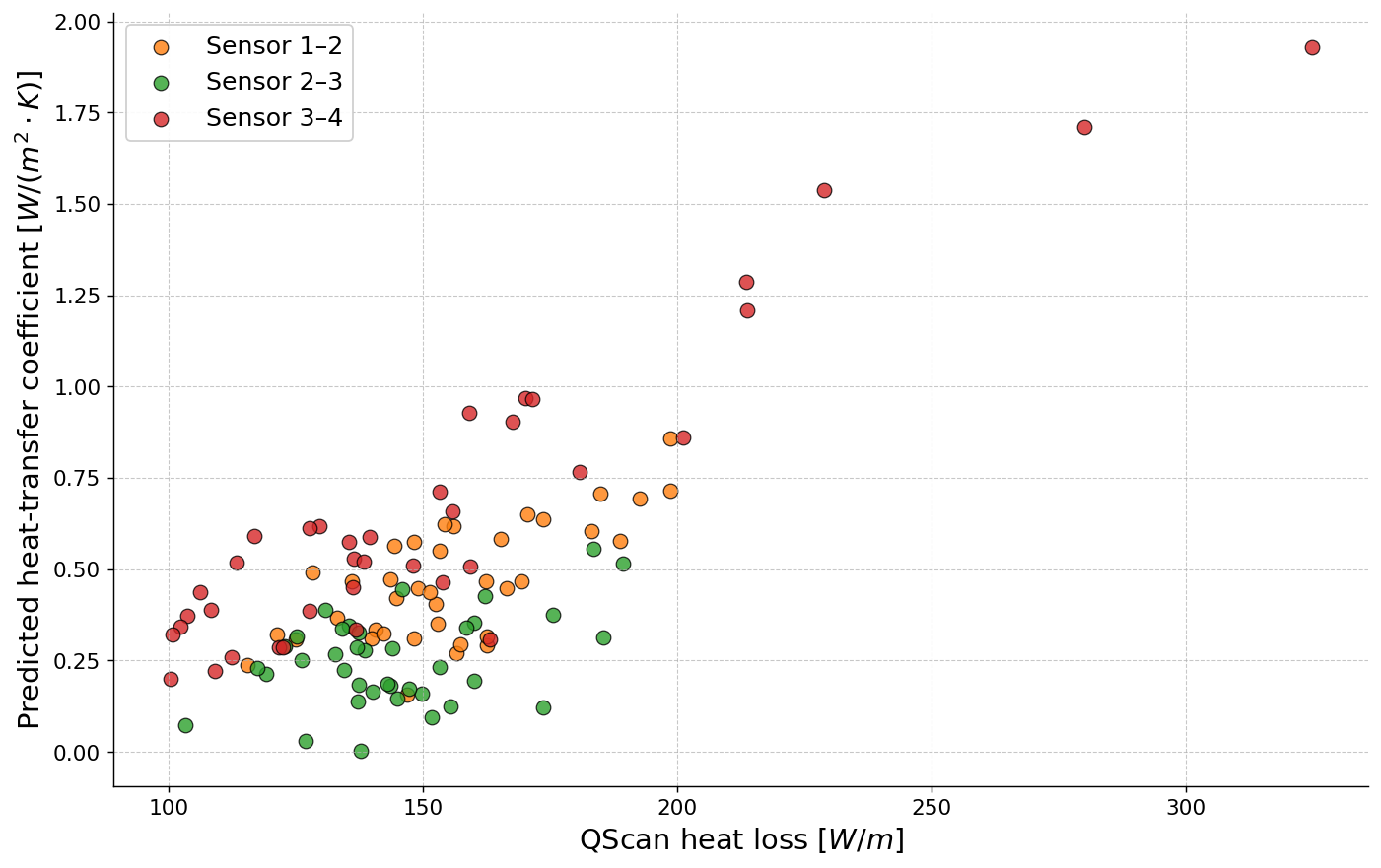}
    \end{subfigure}
    \caption{Left: Drone thermography of receiver heat losses. Right: Predicted $h_{pg}^t$ vs.\ QScan aggregated losses.}
    \label{fig:heat_loss}
\end{figure}

To obtain a ground truth for receiver heat losses, we compare our inferred $h_{pg}^t$ with QScan measurements. QScan is an in-house  CSP Services technology for drone-based solar field surveys (\citet{cspservicesqscan}). The measured temperature results were converted to equivalent nighttime conditions through physical models. Figure~\ref{fig:heat_loss} demonstrates strong qualitative and quantitative agreement, with the exception of the sensor 2--3 region. This discrepancy is most likely attributable to the fact that the drone measurements captured only heat losses at receivers and excluded the better-insulated crossover pipes between sensors 2 and 3.

All areas with high-loss receivers are consistently detected by our algorithm, showing that nighttime operational data contain sufficient signal to identify vacuum degradation.

\section{Conclusion}
We introduced a physics-informed learning framework that infers loop-level mass flows and receiver heat-loss coefficients directly from routine nighttime data in a parabolic trough solar field. By embedding a conjugate heat-transfer model into a differentiable optimization pipeline, the approach recovers latent physical parameters with no need for additional sensors. 

The method achieves accurate temperature predictions, produces stable and interpretable flow distributions, and shows good agreement with drone-based heat-loss measurements. This demonstrates that operational CSP data—when combined with appropriate modeling—enable actionable diagnostics of hydraulic imbalance and receiver degradation. The resulting insights support targeted maintenance and improved field balancing, offering a scalable pathway toward data-enabled performance optimization in commercial CSP plants.


\subsubsection*{Acknowledgments}
We gratefully acknowledge Marquesado Solar for providing the data and valuable technical support. We also thank the German Federal Ministry for the Environment, Nature Conservation, Nuclear Safety and Consumer Protection (BMUV) for funding this work under the AuSeSol‑AI project (grant number: 67KI21007$\%$).

\begin{center}
    \includegraphics[width=0.3\textwidth]{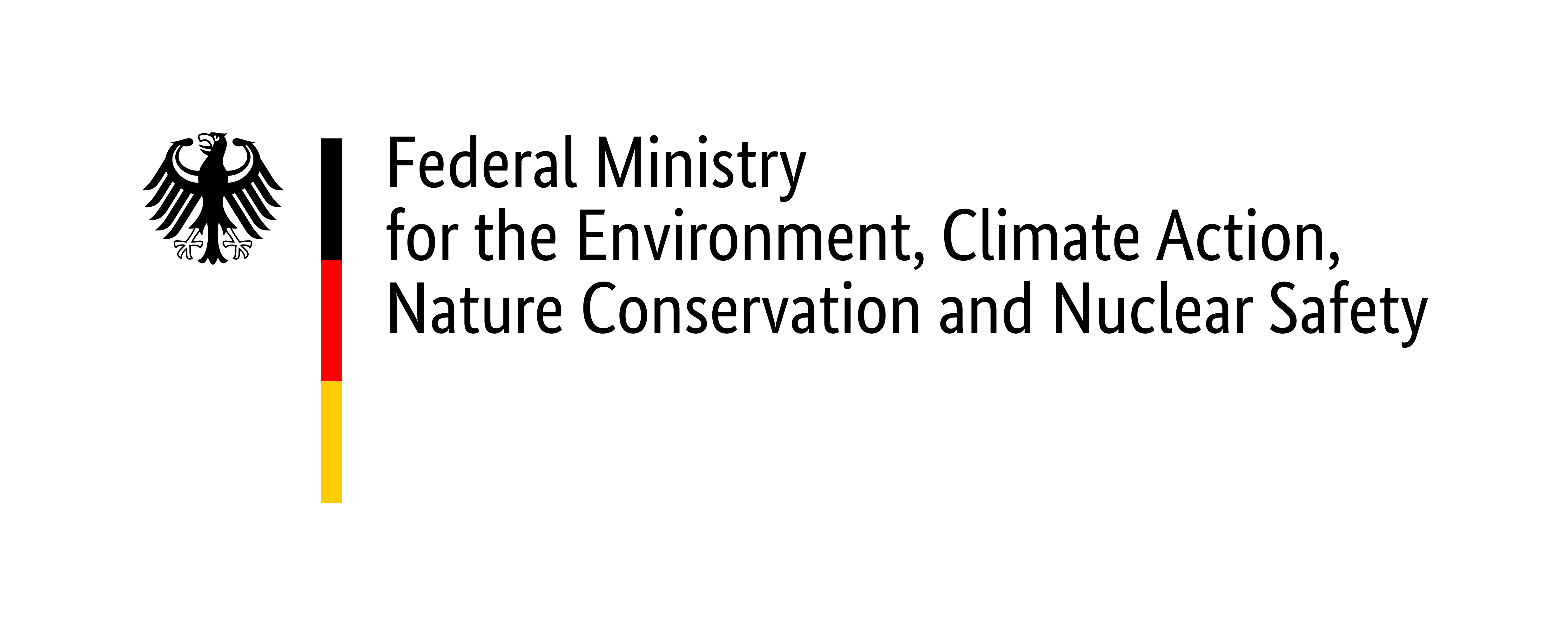}
\end{center}

\bibliography{arxiv_ausesol}
\bibliographystyle{arxiv_ausesol}

\end{document}